\documentclass[sigconf]{acmart}
\usepackage[capitalize]{cleveref}
\usepackage{graphicx}
\usepackage{subcaption}
\usepackage{multirow}
\usepackage{diagbox}
\usepackage{comment}
\AtBeginDocument{%
  }

\setcopyright{acmlicensed}
\copyrightyear{2024}
\acmYear{2024}
\acmDOI{XXXXXXX.XXXXXXX}

\acmConference[MM '24]{Proceedings of the 32th ACM International Conference on Multimedia}{28 October - 1 November 2024}{Melbourne, Australia
}

\acmISBN{978-1-4503-XXXX-X/18/06}

\acmSubmissionID{123}

\begin{document}

\title{Beyond Night Visibility: Adaptive Multi-Scale Fusion of 
Infrared and Visible Images}
%\settopmatter{printacmref=false}
%\renewcommand\footnotetextcopyrightpermission[1]{}
\author{Shufan Pei}
\affiliation{%
  \institution{Fuzhou University}
  \country{}
  }
\email{peishufan@gamil.com}

\author{Junhong Lin}
\affiliation{%
  \institution{Fuzhou University}
  \country{}
  }
\email{jhlin_study@163.com}

\author{Wenxi Liu}
\affiliation{%
  \institution{Fuzhou University}
  \country{}
  }
\email{wenxi.liu@hotmail.com}

\author{Tiesong Zhao}
\affiliation{%
  \institution{Fuzhou University}
  \country{}
  }
\email{t.zhao@fzu.edu.cn}

\author{Chia-Wen Lin}
\affiliation{%
  \institution{National Tsing Hua University}
  \country{}
  }
\email{cwlin@ee.nthu.edu.tw}


\begin{abstract}
  In addition to low light, night images suffer degradation from light effects (e.g., glare, floodlight, etc). However, existing nighttime visibility enhancement methods generally focus on low-light regions, which neglects, or even amplifies the light effects. To address this issue, we propose an Adaptive Multi-scale Fusion network (AMFusion) with infrared and visible images, which designs fusion rules according to different illumination regions. First, we separately fuse spatial and semantic features from infrared and visible images, where the former are used for the adjustment of light distribution and the latter are used for the improvement of detection accuracy. Thereby, we obtain an image free of low light and light effects, which improves the performance of nighttime object detection. Second, we utilize detection features extracted by a pre-trained backbone that guide the fusion of semantic features. Hereby, we design a Detection-guided Semantic Fusion Module (DSFM) to bridge the domain gap between detection and semantic features. Third, we propose a new illumination loss to constrain fusion image with normal light intensity. Experimental results demonstrate the superiority of AMFusion with better visual quality and detection accuracy. The source code will be released after the peer review process.
\end{abstract}

\begin{CCSXML}
<ccs2012>
<concept>
<concept_id>10010147.10010178.10010224</concept_id>
<concept_desc>Computing methodologies~Computer vision</concept_desc>
<concept_significance>500</concept_significance>
</concept>
</ccs2012>
\end{CCSXML}

\ccsdesc[500]{Computing methodologies~Computer vision}

\keywords{Night visibility enhancement, Light effects suppression, Multi-modality fusion, Adaptive multi-scale fusion}


\maketitle

%------------------------------------------------------------------------
\section{Introduction}
\label{sec:intro}
\begin{figure}[t]
    \centering

    \includegraphics[scale=0.65]{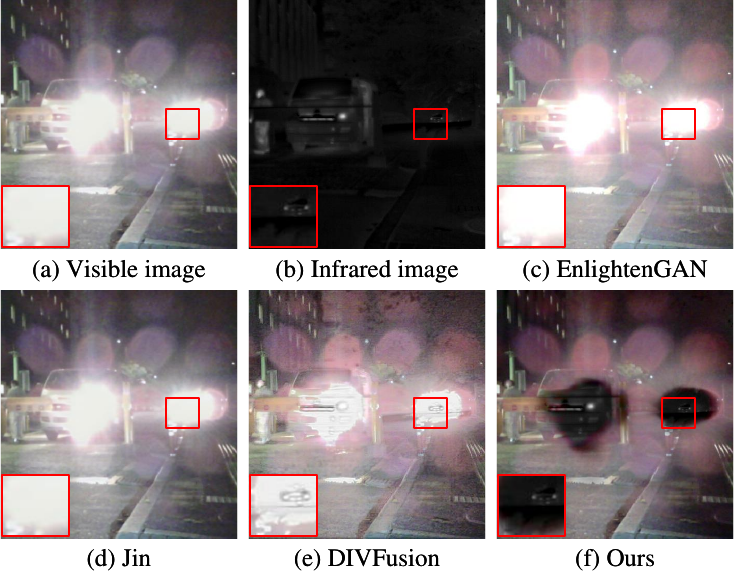}

    \caption{Our method can better remove the masking effect from high beam. (a) Visible image. (b) Infrared image. (c) Low-light enhancement of (a) by \cite{jiang2021enlightengan}. (d) Light-effects suppression of (a) by \cite{jin2022unsupervised}. (e) Multi-modality image fusion result of (a) and (b) by \cite{tang2023divfusion}. (f) Fusion result of (a) and (b) by our method AMFusion.}
    \label{fig:overview}
\end{figure}

Usually, low contrast and severe noise caused by low light are the main problems of nighttime images. However, the over-exposure caused by light effects also needs to be considered, which is a key factor of traffic accidents. The uneven light distributions are very common due to various light effects \cite{sharma2021nighttime} (mainly headlights), as shown in \cref{fig:overview}(a). Such visual quality degradation can further affect the performance of downstream computer vision tasks (e.g., object detection \cite{wang2023yolov7}). Existing nighttime visibility enhancement methods~\cite{guo2016lime, jiang2021enlightengan, li2021learning, zhang2021beyond} are usually focused on improving the visibility of low-light regions, which would, however, inevitably boost the intensity of light-effects regions as shown in \cref{fig:overview}(c). A few unsupervised methods~\cite{sharma2021nighttime, jin2022unsupervised} were proposed to suppress light effects. However, \cref{fig:overview}(d) shows that single modality image-based methods still have some limits. In specific, It is highly difficult to suppress light effects based on single modality input without the constraint of ground truth. But it is intractable to collect paired images with and without light effects for supervised learning. 

%------------------------------------------------------------------------
%------------------------------------------------------------------------

Different from single modality image-based methods, multi-source images can provide extra information for better results. In infrared and visible image fusion (IVIF) \cite{ma2019infrared}, infrared sensors are insensitive to illumination, which can highlight salient targets as shown in \cref{fig:overview}(b). By combining visible image with infrared image, we can get high-quality image free of low light and light effects, which can be applied to military surveillance \cite{das2000color}, object detection \cite{zou2023object} and automatic driving \cite{gao2018object}.

So far, considerable methods based on deep learning for IVIF have been proposed \cite{li2018densefuse, li2020nestfuse, li2021different, liu2022target, ma2019fusiongan, ma2020ddcgan, ma2021stdfusionnet, ma2022swinfusion, tang2022ydtr, zhao2023metafusion}. However, most IVIF methods are designed for normal light conditions. Although a few methods \cite{tang2022piafusion, tang2023divfusion} took illumination into consideration, they still neglect or amplify light effects, which may significantly impact the visual quality and detection accuracy, as shown in \cref{fig:overview}(e). Hence, the removal of light effects remains an important issue to deal with. 

To address this issue, we propose a new unsupervised IVIF network termed Adaptive Multi-scale Fusion network (AMFusion), which designs fusion rules according to different illumination regions. To the best of our knowledge, we are the first that utilize IVIF method to address low light and light effects simultaneously. First, we separately extract spatial and semantic features from infrared and visible images for subsequent processing, since spatial features contain more illumination information while semantic features reflect detection information in compressed dimension. Second, for visual quality, we utilize an illumination adjustment module that extracts the illumination information from spatial features to guide infrared features supply the missing details of low-light and light-effects regions. Third, for detection accuracy, we design a Detection-guided Semantic Fusion Module (DSFM) that better integrate the information from detection features to guide the fusion of semantic features. Forth, we combine the fused spatial and semantic features to obtain fusion image with better visual quality and detection performance. Finally, we design a new illumination loss that constrains fusion image to maintain normal light intensity by unsupervised learning.

In summary, this study presents a new IVIF method for nighttime visibility enhancement that further improves detection performance. Experimental results show that our approach outperforms the state-of-the-art methods. The main contributions are summarized as follows:

\begin{itemize}
    \item We first devise a new IVIF method (AMFusion) to enhance nighttime visibility of regions affected by low light and light effects. We separately fuse spatial and semantic features extracted from visible and infrared images to obtain fusion image with high visual quality, which improves detection accuracy.
    \item We introduce detection features from a pre-trained detection backbone to guide the fusion of semantic features. Hereby, we design DSFM to achieve better feature sharing. Thereby, AMFusion concentrates more on detection targets, which better improves the accuracy of detection.
    \item We propose a new illumination loss to constrain the light intensity of fusion image. Extensive experiments on image fusion and object detection demonstrate the feasibility and effectiveness of our proposed method.
\end{itemize}

%------------------------------------------------------------------------
\section{Related Works}

\textbf{Night Visibility Enhancement.} Existing night visibility enhancement methods \cite{zhang2021beyond,li2021learning, jiang2021enlightengan, jiang2023low} are mainly focused on dealing with the low-light degradations. Jiang \textit{et al.} \cite{jiang2021enlightengan} proposed an unsupervised generative adversarial network termed EnlightGAN. Li \textit{et al.} utilized an intuitive non-linear curve mapping to achieve enhancement. Based on Retinex theory, Zhang \textit{et al.} \cite{zhang2021beyond} built a simple yet effective network named KinD++. Although they achieved promising results in low-light regions, their neglect of light effects leads to the boost of light intensity in light-effects regions.

Recently, a few methods attempted to deal with light effects. Sharma and Tan \cite{sharma2021nighttime} first built a network based on camera response function (CRF) estimation and frequency domain operation to suppress light effects. Jin \textit{et al.} \cite{jin2022unsupervised} introduced an unsupervised method based on layer decomposition to remove light effects. However, the single modality input limits the performance of suppression.

\textbf{Image Fusion.} With the development of deep learning, early works \cite{li2018densefuse, ma2019fusiongan, li2020nestfuse, ma2020ddcgan} applied deep learning-based network to image fusion tasks. Li and Wu first \cite{li2018densefuse} built a deep learning model named DenseFuse for IVIF problem. From then on, more and more learning-based methods were proposed \cite{ma2021stdfusionnet, tang2022ydtr, xu2020u2fusion, li2021different, lin2023ldrm}. Ma \textit{et al.} \cite{ma2022swinfusion} presented an IVIF network based on cross-domain long-range learning and Swin Transformer, termed as SwinFusion. Tang \textit{et al.} \cite{tang2022piafusion} took illumination into account and proposed PIAFusion. Furthermore, they presented a darkness-free IVIF method called DIVFusion \cite{tang2023divfusion} to improve the nighttime fusion performance. Recently, some fusion methods \cite{tang2022image, liu2022target, sun2022detfusion, tang2023rethinking, zhao2023metafusion} paid attention to the performance of downstream tasks besides visual quality. Liu \textit{et al.} \cite{liu2022target} proposed a bilevel optimization formulation and built a target-aware Dual Adversarial Learning called TarDAL to solve the joint problem of fusion and detection. Zhao \textit{et al.} \cite{zhao2023metafusion} presented an IVIF method via meta-feature embedding from object detection to bridge the gap between different tasks.

Although these methods have achieved good results, most of them neglect the influence of the illumination. Though PIAFusion takes illumination into account, its illumination aware module is too simple to adjust complex environment. DIVFusion can light up the darkness in visible image while failing to deal with light effects.

\textbf{Object Dectection}. With the rapid development of deep learning techniques, object detection has made great progress \cite{girshick2015fast, ren2017faster, liu2016ssd, xu2022revisiting, wang2023yolov7, zhang2022group, girshick2014rich}. Joseph \textit{et al.} \cite{redmon2016you} presented YOLO network to predict bounding boxes and class probabilities from full image in one step. Carion \textit{et al.} \cite{carion2020end} first proposed Transformer-based detector termed DETR. Chen \textit{et al.} \cite{chen2023diffusiondet} presented DiffusionDet to formulate object detection as a process from noisy boxes to object boxes.

Most of object detection methods select visible image as signle input. However, the visual quality of visible image can be severely affected by poor illumination conditions  (e.g. low light or light effects), which further influence the accuracy of object detection. By contrast, we introduce infrared image that provides thermal information to improve the performance of object detection.

%------------------------------------------------------------------------
\section{Methodology}
\label{sec3}
\subsection{Problem formulation}\label{sec3.1}
Existing night visibility enhancement methods are usually focused on low-light enhancement by supervised learning, which leads to two main problems: 1) the neglect of light effects limits their performance on visual quality and follow-up detection tasks. 2) it is intractable to collect paired night images with and without light effects for supervised learning \cite{jin2022unsupervised}. Previous SOTA method \cite{jin2022unsupervised} processes low light $L$ and light effects $S$ simultaneously by following the degradation model:
\begin{equation}
    D = O\odot{L} + S,
\label{degradation}
\end{equation}
where $\odot$ represents element-wise multiplication, $D$ represents degraded image and $O$ represents restored image. It decomposes $L$ and $S$ without ground truth, but this unsupervised single input method is highly difficult to eliminate light effects with better visual quality. As shown in the upper part of \cref{fig:method}, it fails when light effects occlude targets. By contrast, we aim to explore a prior image $P$ unaffected by illumination conditions for better constraints. Hereby, we formulate a fusion model ${R}_{P}(\cdot)$ to achieve our goal:
\begin{equation}
    O'=\mathcal{R}_{P}(D,P;\theta)= w_{D} \odot{D} + w_{P} \odot{P},
\label{equation: problem formulation}
\end{equation}
where $\theta$ represents the parameters of $\mathcal{R}_{P}(\cdot)$, $O'$ represents the fusion image, $w_{D}$ and $w_{P}$ represent the weights of visible and prior images, respectively.

\subsection{Infrared and Visible Image Fusion}\label{sec3.2}
\begin{figure}[t]
    \centering
    \includegraphics[scale=0.63]{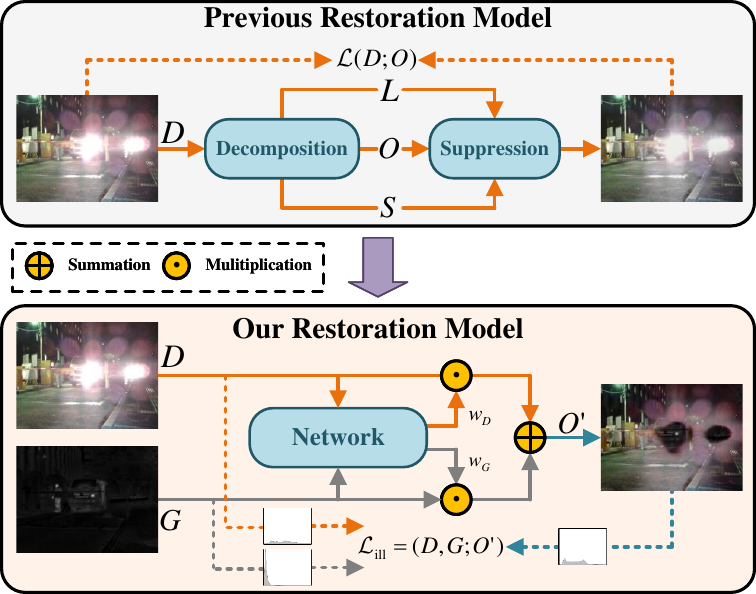}
    \caption{We introduce infrared image to provide extra information, then utilize a fusion model to achieve our goal. Moreover, we design a new illumination loss for normal light distribution.}
    \label{fig:method}
\end{figure}

Different from visible image, infrared image $G$ captures the thermal information and highlight targets, which is insensitive to illumination changes. Therefore, we use infrared image as prior image that provides thermal information to restrain low light and light effects (the lower part of \cref{fig:method}). We rewrite \cref{equation: problem formulation} as:
\begin{equation}
    O'= \mathcal{R}_{F}(D,G;\theta) = w_{D} \odot{D} + w_{G} \odot{G},
\label{equation: fusion}
\end{equation}
where $w_{G}$ represents the weights of infrared image. We aim at using infrared image to supply the loss information of low-light and light-effects regions which exists in visible image with low or high pixel intensity. Moreover, we utilize infrared image to highlight salient targets which are represented as regions with high pixel intensity. Hereby, we use a network $\phi(\cdot)$ to generate the weights:
\begin{equation}
   \left\{w_{D},w_{G}\right\} = \phi(I_{D}(L,S);I_{G}),
\end{equation}
where $I_{D}$ and $I_{G}$ represent the pixel intensity of visible and infrared images, respectively. Considering that spatial features contain more illumination information, we design a network $\mathcal{M}(\cdot)$ to realize \cref{equation: fusion} at feature level for better visual quality. $D$ and $G$ are input to a network $\mathcal{M}(\cdot)$ for feature extraction. We utilize $L$ and $S$ from shallow spatial features $F^{sp}$ to generate fusion features:
\begin{equation}
\begin{aligned}
    &\left\{w_{D},w_{G}\right\} = \phi(F^{sp}_{D}(L,S);F^{sp}_{G}),
    \\
    &F^{sp}_{fu} = w_{D} \odot{F^{sp}_{D}} + w_{G} \odot{F^{sp}_{G}},  
\end{aligned}
\label{equation: featrue fusion}
\end{equation}
where $F^{sp}_{fu}, F^{sp}_{D}$ and $F^{sp}_{G}$ represent fusion, visible and infrared spatial features, respectively. Here, $w_{D}$ in low-light and light-effects regions has a low value to suppress degradation. Accordingly, $w_{G}$ supplies the loss information with high value. Besides, $w_{G}$ of salient targets has a high value to provide extra information.
%-------------------------------------------------------------------------
%-------------------------------------------------------------------------

Besides spatial features, deep semantic features $F^{se}$ reflect detection information in compressed dimension. For the improvement of detection performance, we introduce detection features $F_{d}$ from pre-trained detection backbone $\mathcal{D}$ and  that guide the semantic fusion. However, \cite{zhao2023metafusion} shows there exists task-level differences between fusion and detection. Therefore, we devise a network to bridge the domain gap between detection and fusion features:
\begin{equation}
    F^{se}_{fu} = \psi(F^{se}_{D},F^{se}_{G};F_{d}),
\end{equation}
where $\psi(\cdot)$ denotes a feature sharing network. Finally, we restore $O'$ with spatial and semantic features by a reconstruction network $\mathcal{H}(\cdot)$:
\begin{equation}
    O' = \mathcal{H}(F^{sp}_{fu}, F^{se}_{fu}).
\end{equation}
Hereby, we aim to design $\mathcal{M}(\cdot)$, $\phi(\cdot)$, $\psi(\cdot)$, $\mathcal{H}(\cdot)$ for $\mathcal{R}_{F}(\cdot)$. 

\subsection{Unsupervised Learning}\label{sec3.3}

%-------------------------------------------------------------------------
\begin{figure*}[t]
  \centering
  \includegraphics[scale=0.6]{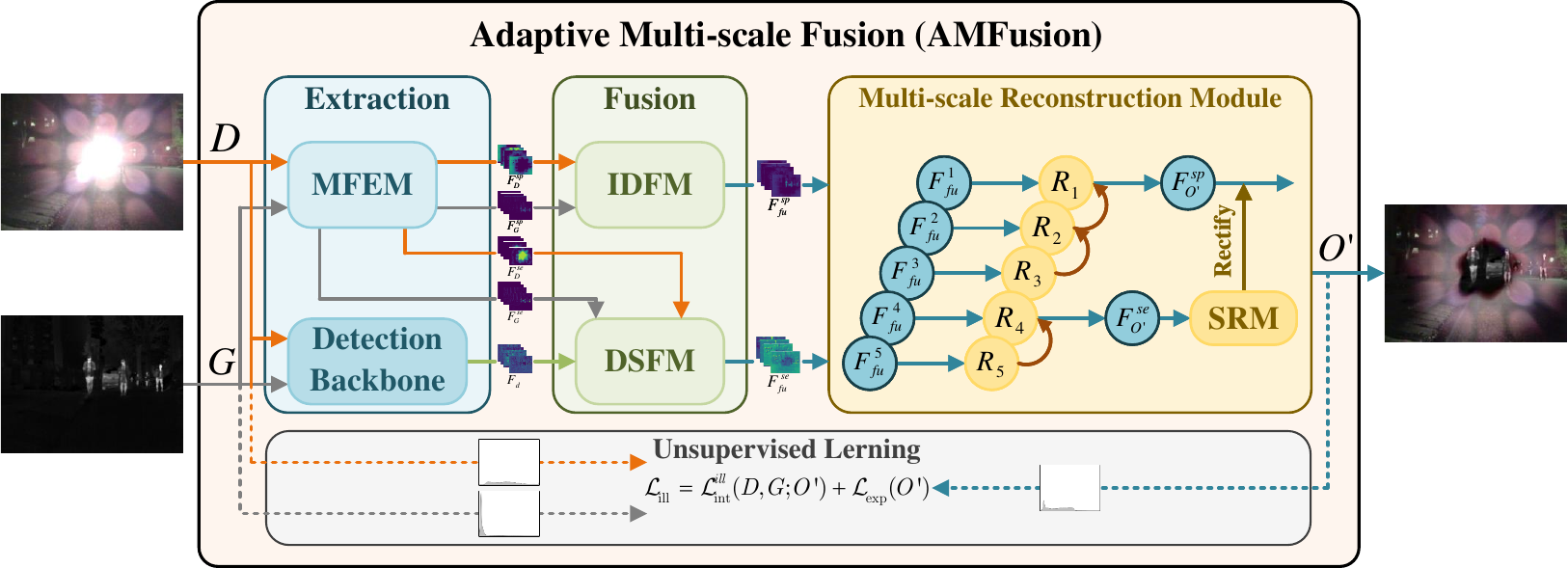}
  \caption{The architecture of our proposed AMFusion. It consists of Multi-scale Feature Extraction Module (MFEM), Illumination-guided Detail Fusion Module (IDFM), Detection-guided Semantic Fusion Module (DSFM) and Multi-scale Reconstruction Module (MRM). MFEM extracts spatial $F^{sp}_{D/G}$ and semantic features $F^{se}_{D/G}$ with different scales. IDFM integrates spatial features based on the light distribution. DSFM utilizes detection features to guide the fusion of semantic features. The details of IDFM and DSFM will be illustrated in \cref{fig:network details}. MRM combines features with all scales to generate fusion image, where $R_{i}$ represents different scale reconstruction module that contains convolution operations and Semantic-Guided Rectify Module (SRM) implements the final adjustment.}
  \label{fig:network}
\end{figure*}
%------------------------------------------------------------------------

In \cref{sec3.2}, we introduce infrared image as new constraint, then utilize an unsupervised fusion model to achieve our goal. Besides model design, loss function is important to constrain model for promising results in unsupervised learning. Previous fusion methods neglect the role of illumination conditions in loss function, which leads to poor training performance. To address this issue, we design a new illumination loss for normal light distribution.

\textbf{Illumination Loss.} In previous fusion methods, the intensity loss $\mathcal{L}_{\mathrm{int}}$ is utilized to maintain the optimal intensity distribution for fusion image, which can be expressed as:
\begin{equation}
    \mathcal{L}_{\mathrm{int}} = \frac{1}{HW}\lVert I_{O'}-\mathrm{max}(I_{D}, I_{G}) \lVert,
\end{equation}
where $H$ and $W$ are the width and height of input image, respectively. $||\cdot||$ represents euclidean norm. However, light-effects regions in visible image usually exist with high pixel intensity, which affects the constraint ability of $\mathcal{L}_{\mathrm{int}}$. In particular, light effects will be preserved. To deal with this problem, we introduce an illumination weight to suppress light effects, which can be written as:
\begin{equation}
w_{\mathrm{ill}}(x)=\frac{1}{3}\sum_{c\in(r,g,b)}\mathrm{exp}(\frac{-(I_{D}^{c}(x)-0.5)^2}{2\sigma^{2}}),
\end{equation}
where $c$ denotes color channel and $\sigma$ measures the exposure quality of one pixel, where we set $\sigma$ as 0.2. Note that input visible images have the intensity range with $[0,1]$ and 0.5 stands for the median of the intensity range. The weight will have a low value if pixel intensity is either low (e.g. a low-light pixel) or high (e.g. a light-effects pixel). After that, we can get a new visible image less affected by light effects, which can be directly fed into the intensity loss. The new illumination-based intensity loss can be formulated as: 
\begin{equation}
    \mathcal{L}_{\mathrm{int}}^{ill} = \frac{1}{HW}\lVert I_{O'}-\mathrm{max}(w_{\mathrm{ill}}\odot I_{D}, I_{G}) \lVert.
\end{equation}
Besides, we introduce an exposure control loss $\mathcal{L}_{\mathrm{exp}}$ \cite{li2021learning} to further restrain low light and light effects. Totally, we obtain the illumination loss:
\begin{equation}
    \mathcal{L}_{\mathrm{ill}} = \mathcal{L}_{\mathrm{int}}^{ill} + \eta \mathcal{L}_{\mathrm{exp}},
\label{equation: ill loss}
\end{equation} 
where $\eta=0.75$ in this paper.

%------------------------------------------------------------------------
\section{Network architecture}
\label{sec4}
\subsection{The Overall Framework}\label{sec4.1}

As shown in \cref{fig:network}, we propose an Adaptive Multi-scale Fusion model (AMFusion) to implement \cref{equation: fusion}. First, Multi-scale Feature Extraction Module (MFEM) generates spatial features $F^{sp}$ and semantic features $F^{se}$ from input images  with different scales, thus  $\mathcal{M}(\cdot)$ is replaced by $\mathcal{M}_{E}(\cdot)$. Detection backbone of YOLOv5s generates detection features, thus $\mathcal{D}$ is replaced by $\mathcal{D}_{Y}$. Second, illumination-guided Detail Fusion Module (IDFM) fuses spatial features and Detection-guided Semantic Fusion Module (DSFM) utilizes detection features $F_{d}$ that guide the fusion of semantic features. Here, we use \cref{equation: featrue fusion} to guide the design of IDFM. Thus, $\phi(\cdot)$ and $\psi(\cdot)$ are replaced by $\phi_{I}(\cdot)$ and $\psi_{D}(\cdot)$, respectively. Finally, Multi-scale Reconstruction Module (MRM) integrates fusion features to reconstruct fusion image, where Semantic-Guided Rectify Module (SRM) uses semantic features to rectify spatial features. Thus, $\mathcal{H}(\cdot)$ is replaced by $\mathcal{H}_{R}(\cdot)$. This process can be expressed as:
\begin{equation}
    \begin{aligned}
        &\left\{F^{sp}_{D/G}, F^{se}_{D/G} \right\} = \mathcal{M}_{E}(D,G),
        \\
        &F^{sp}_{fu} = \phi_{I}(F^{sp}_{D}, F^{sp}_{G}),
        \\
        &F_{d} = \mathcal{D}_{Y}(D,G),
        \\
        &F^{se}_{fu} = \psi(F^{se}_{D},F^{se}_{G};F_{d}),
        \\
        &O'= \mathcal{H}_{R}(F^{sp}_{fu}, F^{se}_{fu}).
    \end{aligned} 
\end{equation}

\subsection{Adaptive Multi-scale Fusion Model}\label{sec4.2}
\textbf{Multi-scale Feature Extraction Module (MFEM).} In order to maintain spatial features with high resolution and preserve semantic features with low resolution, we employ MFEM as a feature extraction network. In MFEM, we send visible and infrared images into two parallel extraction branches, which consist of convolution blocks, dense blocks and down-sampling operations. After that, we obtain features $F^{i}_{D/G}$ with different resolutions. In specific, when $i = 1, 2, 3$, $F^{i}_{D/G}$ represent spatial features $F^{sp}_{D/G}$. If $i = 4, 5$, $F^{i}_{D/G}$ represent features $F^{se}_{D/G}$.

\begin{figure}[t]
  \centering
  \includegraphics[scale=0.73]{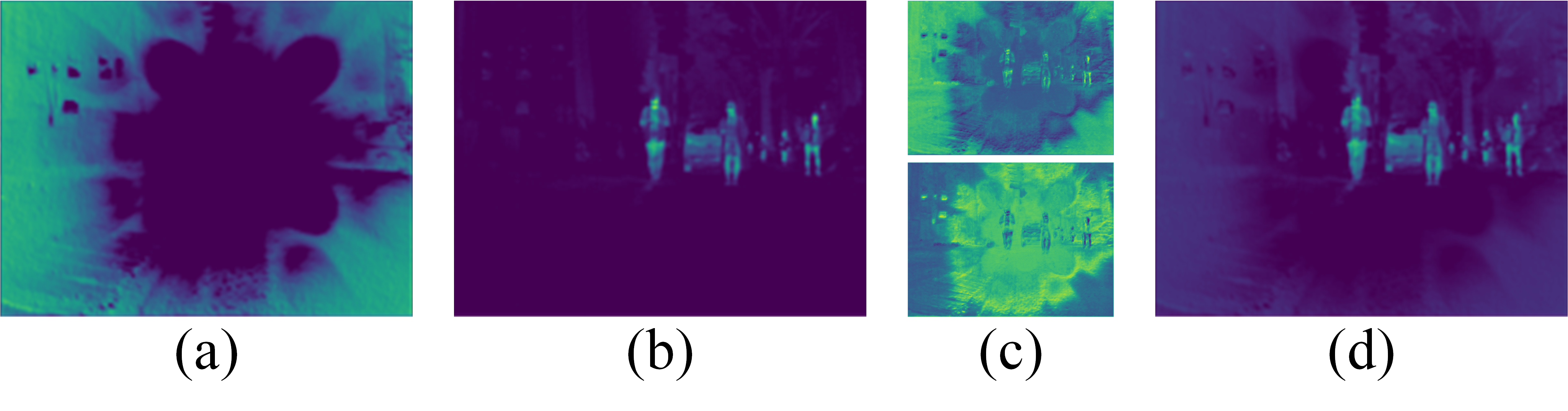}
   \caption{Intermediate results. Brighter pixel represents higher value. (a) Visible features; (b) Infrared features; (c) Weights of a \&b; (d) Fused features.}
   \label{fig:rebuttal1}
   \vspace{-1em}
\end{figure}

\begin{figure}[t]
  \centering
  \begin{subfigure}{1.0\linewidth}
  \includegraphics[scale=0.75]{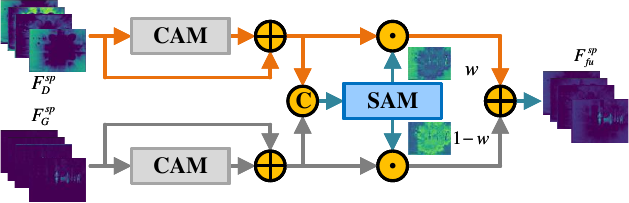}
  \caption{The architecture of IDFM}
  \label{fig:IDFM}
  \end{subfigure}
  \begin{subfigure}{1.0\linewidth}
  \includegraphics[scale=0.7]{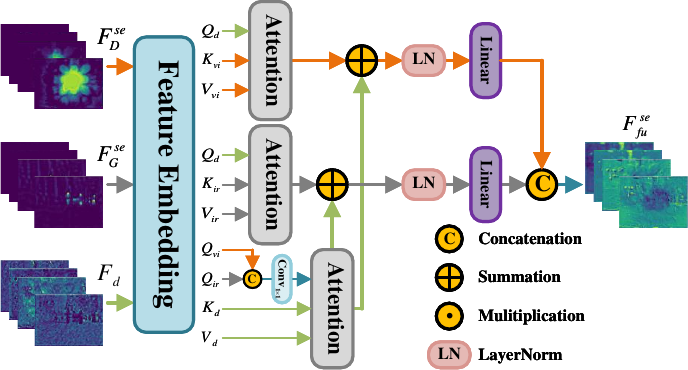}
  \caption{The architecture of DSFM}
  \label{fig:DSFM}
  \end{subfigure}
  \caption{The detailed architecture of IDFM and DSFM.}
  \label{fig:network details}
\end{figure}

\textbf{Illumination-guided Detail Fusion Module (IDFM).} To fully utilize the abundant details and structural information from spatial features, we use IDFM to integrate spatial features. It is based on the channel-spatial attention mechanism \cite{woo2018cbam}, which is shown in \cref{fig:IDFM}. First, spatial features are fed into Channel Attention Module (CAM) to get reinforced features $F_{D}^{c}$ and $ F_{G}^{c}$. Afterward, the reinforced features are concatenated together and fed into Spatial Attention Module (SAM) to get fusion weights $w$. Since visible and infrared spatial features are complementary, we use $w$ as the weights of visible features and $1-w$ as the weights of infrared features. We formulate the final fusion as:
\begin{equation}
    \label{CAFM_fuseq}
    F_{fu}^{sp} = w \odot (F_{D}^{c}+F_{D}^{sp}) +(1-w) \odot (F_{G}^{c}+F_{G}^{sp}).
\end{equation}
In SAM, we follow \cref{equation: featrue fusion} that utilizes the illumination distribution to guide fusion. The region with high or low light intensity in visible image has a low value of weight, thus we get sufficient spatial information without being affected by low light and light effects. To achieve this goal, we design a new illumination loss that constrains the fusion result, which has been illustrated in \cref{sec3.3}. \cref{fig:rebuttal1} enlarges the intermediate results in \cref{fig:network}. We can see adaptive weights successfully fuse both infrared and visible features, which demonstrates the feasibility of our strategy.

%------------------------------------------------------------------------
%------------------------------------------------------------------------

\textbf{Detection-guided Semantic Fusion Module (DSFM).}  We utilize the cross-attention mechanism to implement the fusion of semantic features, as shown in \cref{fig:DSFM}. Considering that semantic features have low resolutions, the introduction of cross-attention will not bring much computational cost. Besides, we introduce detection features $F_{d}$ to guide the fusion. We feed $F_{D}^{se}$, $F_{G}^{se}$ and $F_{d}$ into the embedding model to get embedding features $F_{D}^{e}$, $F_{G}^{e}$ and $F_{d}^{e}$. Then we divide embedding features into Query, Key and Value$\left\{Q_{x}, K_{x}, V_{x}\right\}\in\mathbb{R}^{HW\times{C}}$, where $x\in\left\{D, G, d\right\}$. $H$, $W$ and ${C}$ represent the size of input semantic features. $Q_{d}$ denotes the semantic features that detection model focuses on. $Q_{D}$ and $Q_{G}$ represent those concerned by fusion model. The cross-attention process can be expressed as:
\begin{equation}
\begin{aligned}
    &\left\{F_{D}^{d}, F_{G}^{d}\right\}=\left\{\mathrm{attn}(Q_{d}, K_{D}, V_{D}), \mathrm{attn}(Q_{d}, K_{G}, V_{G})\right\},
    \\
    &F_{d}^{f}=\mathrm{attn}(\mathrm{Conv}_{1\times{1}}(\mathrm{Concat}(Q_{D},Q_{G})), K_{d}, V_{d}),
    \\
    &F_{D}^{a}=\mathrm{Reshape(Linear(LN(}F_{d}^{f}+F_{D}^{d}))),
    \\
    &F_{G}^{a}=\mathrm{Reshape(Linear(LN(}F_{d}^{f}+F_{G}^{d}))),
    \\
    &F_{fu}^{se}=\mathrm{Conv_{3\times{3}}(Concat(}F_{D}^{a},F_{G}^{a})),
\end{aligned}
\end{equation}
where $\mathrm{attn}(\cdot)$ denotes the common multi-head attention module and $\mathrm{LN(\cdot)}$ represents LayerNorm. With the help of detection features, we can generate the semantic features $F_{fu}^{se}$ that contain more information of targets.
%------------------------------------------------------------------------

%------------------------------------------------------------------------
\textbf{Multi-scale Reconstruction Module (MRM).} After getting fusion features $F_{fu}^{sp/se}$ with different scales from IDFM and DSFM, we utilize MRM module for final reconstruction. As shown in \cref{fig:network}, features with low resolution are sent to convolution and up-scaled to be concatenated with those with high resolution. After the bottom-up fusion process, we can get final spatial and semantic fusion features $F^{sp/se}_{O'}$. To integrate them together, we develop SRM to convert $F^{se}_{O'}$ into semantic weight and bias that rectify $F^{sp}_{O'}$. In particular, it consists of two $3\times3$ convolutions with ReLU activation to get weight and bias separately. The rectified feature is utilized to get final fusion image:
\begin{equation}
\left\{w_{se}, b_{se}\right\} = \mathrm{SRM}(F^{se}_{O'}),
\end{equation}
\begin{equation}
O'= \mathrm{Conv}(w_{se} \odot F^{sp}_{O'} + b_{se}).
\end{equation}

\subsection{Loss function}\label{sec4.3}
\label{loss}
The total loss function $\mathcal{L}_{f}$ consists of three types of losses, \textit{i.e.}, gradient loss $\mathcal{L}_{\mathrm{grad}}$, structure loss $\mathcal{L}_{\mathrm{SSIM}}$ and illumination loss $\mathcal{L}_{\mathrm{ill}}$ in \cref{equation: ill loss}. The gradient loss $\mathcal{L}_{\mathrm{grad}}$ and structure loss $\mathcal{L}_{\mathrm{SSIM}}$ can be expressed as:
\begin{equation}
\mathcal{L}_{\mathrm{grad}} = \frac{1}{HW}\lVert  |\nabla O'|- \mathrm{max}(|\nabla D|,|\nabla G|) \lVert_{1},
\end{equation}
\begin{equation}
\mathcal{L}_{\mathrm{SSIM}}=(1-\mathrm{SSIM}_{O',D})/2 + (1-\mathrm{SSIM}_{O',G})/2,
\end{equation}
where $\nabla$ refers to Sobel gradient operator and $|\cdot|$ denotes absolute operation. $H$ and $W$ are the width and height of input image, respectively. The total fusion loss is:
\begin{equation}
    \mathcal{L}_{f} = \mathcal{L}_{\mathrm{grad}} + \alpha \mathcal{L}_{\mathrm{SSIM}} + \beta \mathcal{L}_{\mathrm{ill}},
\end{equation}
where we set $\alpha=1.5$ and $\beta=2.0$.

%------------------------------------------------------------------------
\section{Experiments}
\subsection{Datasets and Evaluation}

%------------------------------------------------------------------------

\begin{figure*}[!ht]
  \centering
  \includegraphics[scale=0.55]{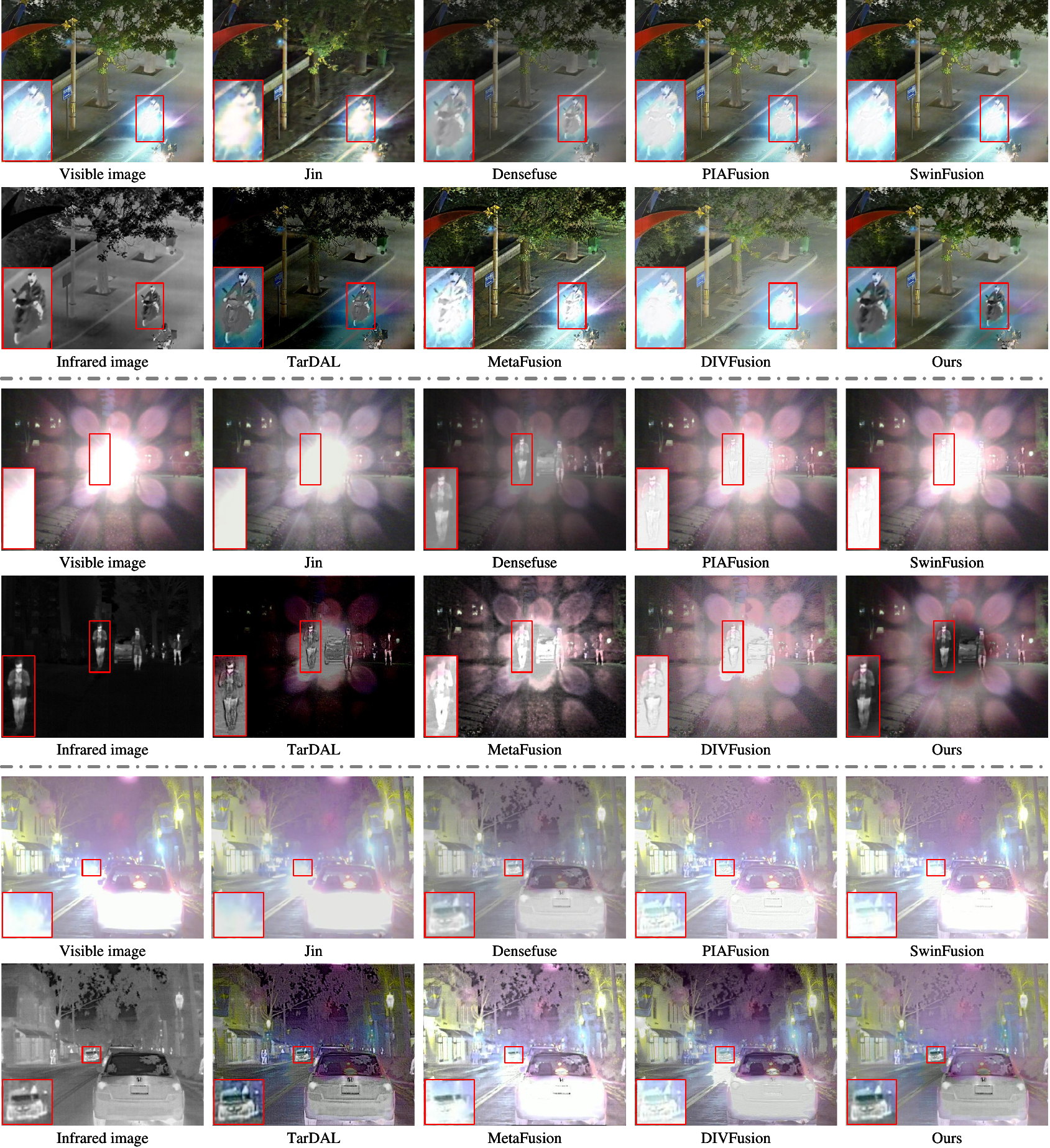}
  \caption{Visual comparisons of different fusion methods on $\mathrm{M^{3}FD}$+LLVIP (first group), MSRS (second group) and Roadscene(third group). The results of other methods can be found in the supplemental material.} 
  \label{fig:fusion qualitative comparisons}
  %\vspace{0.2em}
\end{figure*}

%------------------------------------------------------------------------
\textbf{Datasets.} To satisfy our application scene, we implement qualitative and quantitative experiments on $\mathrm{M^{3}FD}$ \cite{liu2022target} and LLVIP \cite{jia2021llvip} datasets. $\mathrm{M^{3}FD}$ dataset contains well-aligned infrared/visible image pairs with six detection classes. LLVIP dataset consists of aligned infrared/visible night images all collected in the street, which also contains numerous annotated objects. In specific, we select 2100/200 pairs of night images from $\mathrm{M^{3}FD}$ and LLVIP datasets as training set and testing set (900 images from $\mathrm{M^{3}FD}$ and 1400 images from LLVIP). Here we name the testing set as $\mathrm{M^{3}FD}$+LLVIP. Besides, we choose nighttime images from MSRS \cite{tang2022piafusion} and Roadscene \cite{xu2020u2fusion} to demonstrate the generalization ability of our method. We select 90 typical images from MSRS and 20 representative images from Roadscene for generalization analysis.

%------------------------------------------------------------------------
\begin{table*}[ht]\renewcommand\arraystretch{1.4}
    \caption{Quantitative comparisons of different methods on $\mathrm{M^{3}FD}$+LLVIP, MSRS and Roadscene. The best result is in \textcolor{red}{red} and the second best one is in \textcolor{blue}{blue}.}
    \centering
    \tabcolsep=0.15cm
    \begin{tabular}{@{}c|ccc|ccc|ccc|c@{}}
\toprule
                         & \multicolumn{3}{c|}{$\mathrm{M^{3}FD}$+LLVIP}                                                              & \multicolumn{3}{c|}{MSRS}                                                                    & \multicolumn{3}{c|}{Roadscene}                                                               & Detection                   \\ \cline{2-11} 
\multirow{-2}{*}{Method} & EN                           & MI                            & SD                            & EN                           & MI                            & SD                            & EN                           & MI                            & SD                          & {\footnotesize m$AP_{50\rightarrow95}$(\%)}                      \\ \midrule
EnlightGAN \cite{jiang2021enlightengan} (TIP'21)               & 6.521                             &   \diagbox[dir=SW]{}{}                           & 33.198                              &  5.430                            &\diagbox[dir=SW]{}{}                           &   28.961                            & 6.625                          & \diagbox[dir=SW]{}{}                       &    31.061                           & 51.3                        \\
Jin \cite{jin2022unsupervised} (ECCV'22)                      & 6.224                        & \diagbox[dir=SW]{}{}                              & 33.333                        & 5.791                             & \diagbox[dir=SW]{}{}                              &  30.008                             & 7.027                             & \diagbox[dir=SW]{}{}                              &  38.847                             & 49.4                        \\ \midrule
DenseFuse \cite{li2018densefuse} (TIP'18)                & 6.601                        & 13.202                        & 29.329 & 5.375                        & 10.753                        & 18.848                        & 6.767                        & 13.534                        & 36.677                        & 62.5                        \\
YDTR \cite{tang2022ydtr} (TMM'22)                     & 6.953                        & 13.906                        & 39.249                        &  5.602 & 11.201                        & 25.375 & 6.685                        & 13.370                        & 32.658                        &  62.3 \\
PIAFusion \cite{tang2022piafusion} (InfFus'22)               & 7.134                        & 14.268                        & {\color[HTML]{0500FC} 45.392} & 5.631                        & 11.918                        & 30.359                        & 6.687                        & 13.746                        & 41.418 & 62.3                        \\
SwinFusion \cite{ma2022swinfusion} (IJCAI'22)               & 7.159                        & 14.319                        & 45.261                        & 5.775                        & 11.998                        & {\color[HTML]{0500FC} 32.669} & 6.734                        & 13.468                        & 43.186                        & {\color[HTML]{0500FC} 62.8} \\
TarDAL \cite{liu2022target} (CVPR'22)                  & 4.670 &  9.341  & 38.937                        & 3.285 & 1.414  & 14.254 &  7.110 & 14.221 & \textcolor{red}{49.185} & 62.6                        \\
MetaFusion \cite{zhao2023metafusion} (CVPR'23)              & 6.840 &  13.681 &  42.972 & 5.511 & 11.029 & {\color[HTML]{333333} 30.013} &  7.104 & 14.208 &  46.484 &  60.9 \\
DIVFusion \cite{tang2023divfusion} (InfFus'23)               & {\color[HTML]{0500FC} 7.294} & {\color[HTML]{0500FC} 14.389} & 44.139                        & {\color[HTML]{FF0000} 6.063} & {\color[HTML]{FF0000} 13.801} & {\color[HTML]{FF0000} 35.854} & {\color[HTML]{FF0000} 7.490} & {\color[HTML]{0500FC} 14.282} & {\color[HTML]{0500FC} 48.463} & 57.2                        \\ \hline
Ours                     & {\color[HTML]{FF0000} 7.323} & {\color[HTML]{FF0000} 14.446} & {\color[HTML]{FF0000} 46.032} & {\color[HTML]{0500FC} 5.844} & {\color[HTML]{0500FC} 12.291} & 30.513 & {\color[HTML]{0500FC} 7.177} & {\color[HTML]{FF0000} 14.353} & 43.946                        & {\color[HTML]{FF0000} 63.1} \\ \bottomrule
\end{tabular}

    \label{tab:quantitative comparisons}
    \vspace{0.1em}
\end{table*}
%------------------------------------------------------------------------

\textbf{Evaluation.} We compare the fusion quality with seven fusion SOTA methods based on qualitative and quantitative analyses, including DenseFuse \cite{li2018densefuse}, YDTR \cite{tang2022ydtr}, SwinFusion \cite{ma2022swinfusion}, PIAFusion \cite{tang2022piafusion}, TarDAL \cite{liu2022target}, MetaFusion \cite{zhao2023metafusion} and DIVFusion \cite{tang2023divfusion}. Besides, we select two methods (EnlightenGAN \cite{jiang2021enlightengan} and Jin \cite{jin2022unsupervised}) based on single modality input. Three objective metrics of visual quality are leveraged for the comparison, including entropy (EN) \cite{roberts2008assessment}, mutual information (MI) \cite{qu2002information} and standard deviation (SD) \cite{aslantas2015new}. Besides, we select $\mathrm{mAP_{50\rightarrow95}}$ \cite{he2022destr} for comparisons on detection performance.

\textbf{Implementation.} We select the YOLOv5s trained on the VOC dataset as the pre-trained detection backbone. We do not integrate AMFusion and detection network for training together due to their task-level difference. First, we freeze the parameters of pre-trained backbone to train AMFusion for 200 epochs, where the initial learning rate is set $2\times{10^{-4}}$ and progressively decays to $1\times{10^{-6}}$. Then we unfreeze all parameters to fine tune the model for 50 epochs, where the initial learning rate is set $1\times{10^{-6}}$ and finally decays to $1\times{10^{-8}}$. All steps are optimized by Adam. The input images are randomly cropped to $256\times{256}$ with the batchsize of 16. All experiments are performed on a server equipped with  RTX 3090.

%------------------------------------------------------------------------
\begin{figure}[t]
    \centering
    \includegraphics[scale=0.45]{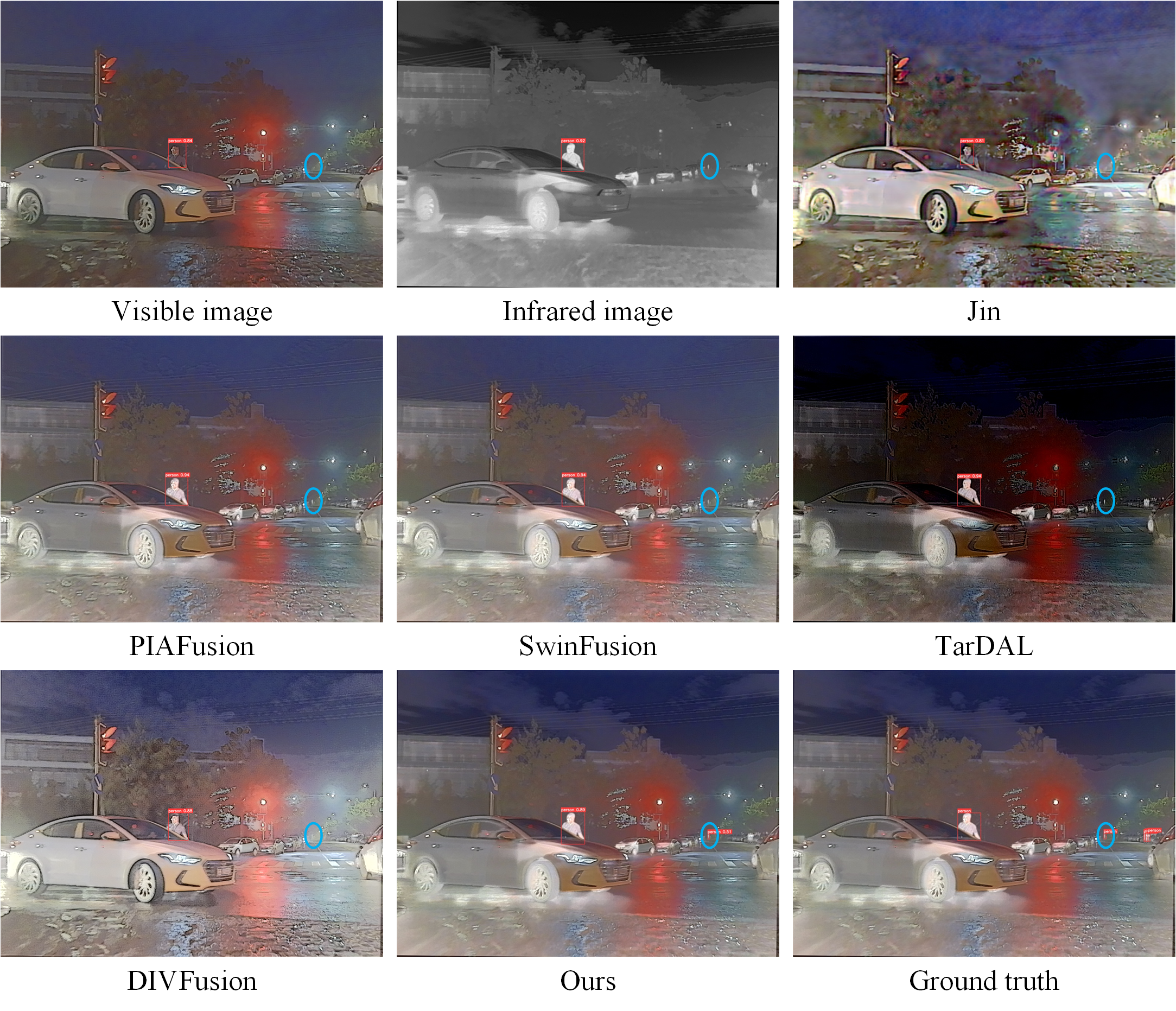}
    \caption{Qualitative comparisons of object detection results between different methods on $\mathrm{M^{3}FD}$+LLVIP. Comparisons of ohter methods can be found in the supplement material.}
    \label{fig:qualtative comparions od}
    \vspace{-0.5em}
\end{figure}
%------------------------------------------------------------------------

\subsection{Comparison on Visual Quality}

\textbf{Qualitative Comparisons}. The qualitative results on $\mathrm{M^{3}FD}$+LLVIP and MSRS are depicted in \cref{fig:fusion qualitative comparisons}. We can see that the single modality image-based method can not suppress light effects. All the fusion methods can extract and fuse main features from infrared and visible images to some extent. However, images generated from DenseFuse are darker than source images, which results from its handicraft fusion rules. MetaFusion produces a lot of noise. Besides, all of them can not handle the light effects well, as shown in the red rectangular boxes. Though TarDAL removes light effects to some extent, it suffers from abnormal light distributions. By contrast, our proposed AMFusion can generate images free of light effects.

\textbf{Quantitative Comparisons.} We plot the numerical results of different methods in \cref{tab:quantitative comparisons}. Since EnlightenGAN and Jin are not image fusion methods, we do not compare the MI values of them. Obviously our method generally achieves outstanding metric values. Specifically, the higher EN indicates that our fused images contain abundant information. And compared with single modality image-based methods, the introduction of infrared images (higher EN) brings better visual quality (better SD), which demonstrates the effectiveness of extra information. The larger value of MI reveals that our method can preserve more considerable information from source images. Moreover, large SD implies that our results have better contrast. It is noted that all methods do not perform well in MSRS dataset due to its extremely low light condition. Besides, DIVFusion gets the promising value in three datasets. It incorporates a low-light enhancement (LLE) module. However, the qualitative result shows that it boosts the intensity of the whole image including light-effects regions, which results the amplification of noise. Too much noise also can lead to higher values of EN and SD. By contrast, our method performs the best on $\mathrm{M^{3}FD}$+LLVIP and the 2nd best on other datasets without LLE module. And qualitative comparisons show that only our method can successfully suppress light effects. In general, our method achieves best results.

\subsection{Comparison on Object Detection}
We evaluate object detection accuracy on infrared images, visible images and images generated from different methods. We utilize YOLOv5s as the baseline method for object detection. For fair comparison, we use the results of different methods to retrain the detection model, respectively. Hereby, we only test the accuracy of pedestrian detection. 

\textbf{Qualitative Comparisons.} We select $\mathrm{M^{3}FD}$+LLVIP for fair comparisons. For clear presentation, we label the bounding boxes of annotated pedestrians on fused images as ground-truths. As shown in \cref{fig:qualtative comparions od}, detection model can not get promising results from single modality image-based methods. By contrast, all the fusion methods can improve the performance of detection by utilizing complementary information from source images. Among them, our proposed method achieves the best result, e.g., the small target is accurately detected (see the blue ellipse).

\textbf{Quantitative Comparisons.} We plot the numerical results of different methods for comprehensively comparisons. As shown in \cref{tab:quantitative comparisons}, our proposed method achieves the highest accuracy. Besides, $\rm{mAP}_{50 \rightarrow 95}$ values with infrared and visible images are 60.8\% and 50.9\%, respectively, which are lower than 63.1\% with AMFusion. In general, AMFusion can produce visual appealing fusion result as well as improve the performance of object detection.

%------------------------------------------------------------------------
\begin{figure}[t]
    \centering
    \includegraphics[scale=0.47]{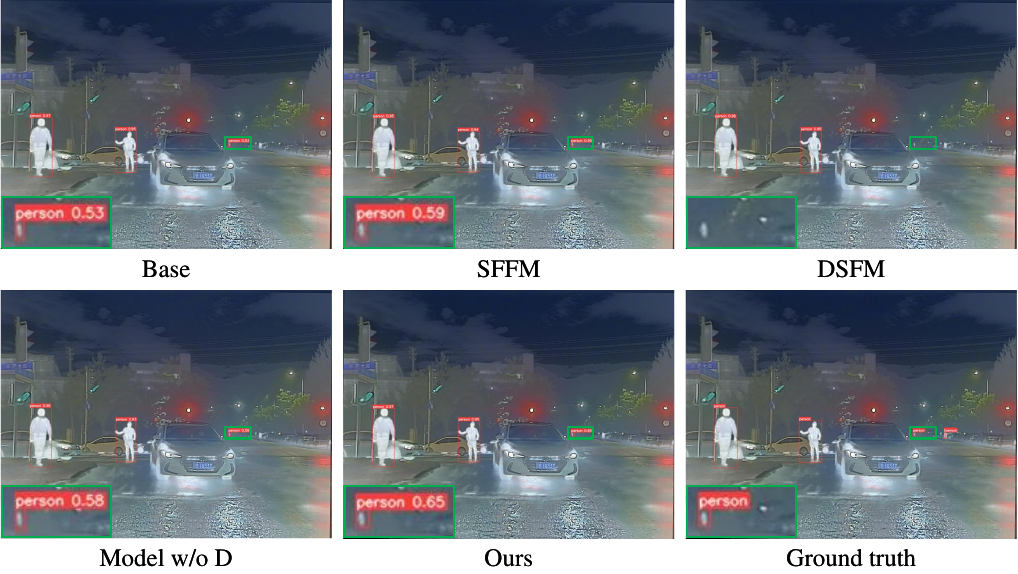}
    \caption{Visual results of different modules on \textbf{$\mathrm{M^{3}FD}$}+LLVIP.}
    \label{fig:ablation modules}
\end{figure}

%------------------------------------------------------------------------
%------------------------------------------------------------------------
\begin{table}[t]\renewcommand\arraystretch{1.3}
\centering
\caption{Quantitative results of different frameworks on $\mathrm{M^{3}FD}$+LLVIP.}
\tabcolsep=0.11cm
\begin{tabular}{@{}ccccc|cc@{}}
\toprule
\textbf{Base} & \textbf{IDFM} & \textbf{DSFM} & \textbf{SRM} & \textbf{w/D} & EN                 & m$AP_{50\rightarrow95}$(\%)           \\ \midrule
\checkmark             &               &               &              & \checkmark            & 7.109                        & 60.4                        \\
\checkmark             & \checkmark             &               &              & \checkmark            & 7.184                        & 60.8                        \\
\checkmark             & \checkmark             & \checkmark             &              & \checkmark            & 7.215                        & 61.6                        \\
\checkmark             & \checkmark             & \checkmark             & \checkmark            &              & 7.211                        & 61.1                        \\ \hline
\textbf{\checkmark}             & \textbf{\checkmark}             & \textbf{\checkmark}             & \textbf{\checkmark}            & \textbf{\checkmark}            & \textbf{7.323} & \textbf{61.7} \\ \bottomrule
\end{tabular}

\label{tab: ablation modules}
\end{table}
%------------------------------------------------------------------------

%------------------------------------------------------------------------
\begin{figure}[t]
    \centering
    \includegraphics[scale=0.45]{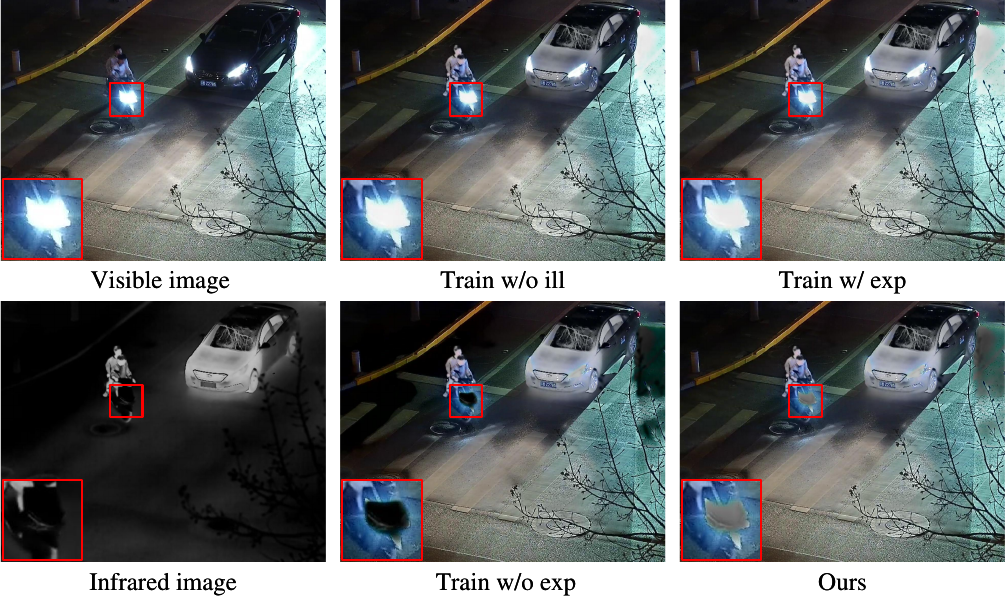}
    \caption{Visual results of different losses on $\mathrm{M^{3}FD}$+LLVIP.}
    \label{fig:ablation loss}
\end{figure}
%------------------------------------------------------------------------

%------------------------------------------------------------------------
\begin{table}[t]\renewcommand\arraystretch{1.3}
\centering
\caption{Influence study of illumination by comparing different loss functions on $\mathrm{M^{3}FD}$+LLVIP.} 
\tabcolsep=0.25cm
\begin{tabular}{c|ccc}
\toprule
Method        & EN & MI & SD \\ \midrule
Train w/o $\mathcal{L}_{\mathrm{ill}}$ &  6.978  &  13.957  &  41.078  \\
Train w/o $\mathcal{L}_{\mathrm{int}}^{ill}$  & 7.174   &  14.347  &  45.302  \\
Train w/o $\mathcal{L}_{\mathrm{exp}}$ &  7.217  &  14.432  & 45.302   \\ \hline
\textbf{Ours}          & \textbf{7.323}   & \textbf{14.446}   & \textbf{46.032}  \\ \bottomrule
\end{tabular}

\label{tab: ablation loss}
\end{table}
%------------------------------------------------------------------------
\subsection{Ablation Study}

\textbf{Effect of Guidance of Detection Features.} In \cref{sec4.2}, we introduce detection features to help the fusion of semantic features. To verify their effects, the model without detection features (w/o D) and our proposed method are compared, where the deep semantic features do cross-attention operation themselves. As shown in \cref{tab: ablation modules} and \cref{fig:ablation modules}, with the help of detection features, we can get fusion images for better visual quality and detection performance.

\textbf{Contributions of Different Modules.} We explore the contributions of three main modules in AMFusion by adding them incrementally. Specially, we explore the effects of \textbf{IDFM}, \textbf{DSFM} and \textbf{SRM}. \textbf{Base} represents AMFusion without using all three modules. Here IDFM is replaced with traditional Convolutional Block Attention Module (CBAM) \cite{woo2018cbam}. DSFM is replaced by self-attention mechanism. SRM is removed and semantic features are directly fused with spatial features. Then we gradually add IDFM, DSFM and SRM to examine their effects on performance. From the results of \cref{tab: ablation modules} and \cref{fig:ablation modules}, each module has made a contribution to performance and our proposed AMFusion has best detection result.

\textbf{Influence of Illumination Loss.} In \cref{sec3.3}, we design an illumination loss $\mathcal{L}_{\mathrm{ill}}$ to constrain the fusion image with normal light distribution, which consists of illumination-based intensity loss $\mathcal{L}_{\mathrm{int}}^{ill}$ and exposure control loss $\mathcal{L}_{\mathrm{exp}}$. To validate their influence, we compare training without illumination loss (Train w/o $\mathcal{L}_{\mathrm{ill}}$), only with exposure loss (Train w/o $\mathcal{L}_{\mathrm{int}}^{ill}$), without exposure loss (Train w/o $\mathcal{L}_{\mathrm{exp}}$) and with all losses (Ours). The quantitative results are shown in \cref{tab: ablation loss}, which illustrate the effectiveness of our illumination loss in visual quality. \cref{fig:ablation loss} shows that each loss has made a contribution to the removal of light effects. Our method that utilizes the full illumination loss to constrain the model achieves best visual results.

%------------------------------------------------------------------------
\section{Conclusion}
This paper presents AMFusion to deal with low light and light effects in night images, which designs fusion rules according to different illumination regions. In AMFusion, spatial and semantic features are separately extracted from input images. Then IDFM utilizes illumination information to guide spatial feature fusion with better visual quality. DSFM utilizes introduced detection features to guide semantic feature fusion with higher detection accuracy. Moreover, a new designed illumination loss effectively constrains fusion images with normal light intensity. Experimental results show that our method generates fusion images with high visual quality and detection accuracy.

%------------------------------------------------------------------------

\bibliographystyle{ACM-Reference-Format}
\bibliography{ref}

\begin{comment}
\appendix

\section{Research Methods}

\subsection{Part One}

Lorem ipsum dolor sit amet, consectetur adipiscing elit. Morbi
malesuada, quam in pulvinar varius, metus nunc fermentum urna, id
sollicitudin purus odio sit amet enim. Aliquam ullamcorper eu ipsum
vel mollis. Curabitur quis dictum nisl. Phasellus vel semper risus, et
lacinia dolor. Integer ultricies commodo sem nec semper.

\subsection{Part Two}

Etiam commodo feugiat nisl pulvinar pellentesque. Etiam auctor sodales
ligula, non varius nibh pulvinar semper. Suspendisse nec lectus non
ipsum convallis congue hendrerit vitae sapien. Donec at laoreet
eros. Vivamus non purus placerat, scelerisque diam eu, cursus
ante. Etiam aliquam tortor auctor efficitur mattis.

\section{Online Resources}

Nam id fermentum dui. Suspendisse sagittis tortor a nulla mollis, in
pulvinar ex pretium. Sed interdum orci quis metus euismod, et sagittis
enim maximus. Vestibulum gravida massa ut felis suscipit
congue. Quisque mattis elit a risus ultrices commodo venenatis eget
dui. Etiam sagittis eleifend elementum.

Nam interdum magna at lectus dignissim, ac dignissim lorem
rhoncus. Maecenas eu arcu ac neque placerat aliquam. Nunc pulvinar
massa et mattis lacinia.
\end{comment}

\end{document}